\documentclass{article}

\usepackage{PRIMEarxiv}

\usepackage[utf8]{inputenc} 
\usepackage[T1]{fontenc}    
\usepackage{hyperref}       
\usepackage{url}            
\usepackage{booktabs}       
\usepackage{amsfonts}       
\usepackage{nicefrac}       
\usepackage{microtype}      
\usepackage{lipsum}
\usepackage{fancyhdr}       
\usepackage{graphicx}       
\usepackage{animate}
\usepackage{placeins}
\usepackage{amsmath}
\usepackage{cleveref}
\usepackage{subfigure}
\crefformat{section}{\S#2#1#3}
\crefformat{subsection}{\S#2#1#3}
\crefformat{subsubsection}{\S#2#1#3}
\crefrangeformat{section}{\S\S#3#1#4 to~#5#2#6}
\crefmultiformat{section}{\S\S#2#1#3}{ and~#2#1#3}{, #2#1#3}{ and~#2#1#3}
\graphicspath{{media/}}     

\title{SineKAN: Kolmogorov-Arnold Networks Using Sinusoidal Activation Functions
\thanks{\textit{\underline{Citation}}: 
\textbf{E.A.F. Reinhardt$^1$, P.R. Dinesh, H.V. Nguyen, S. Gleyzer. SineKAN: Kolmogorov-Arnold Networks Using Sinusoidal Activation Functions.}} 
}

\author{
  Eric A. F. Reinhardt\footnotemark[1] \\
  Department of Physics and Astronomy \\
  The University of Alabama \\
  Tuscaloosa, AL \\
  \texttt{eareinhardt@crimson.ua.edu} \\
   \And
  P. R. Dinesh \\
  Department of Physics and Astronomy \\
  The University of Alabama \\
  Tuscaloosa, AL \\
  \texttt{dpr@crimson.ua.edu} \\
   \And
  Sergei Gleyzer \\
  Department of Physics and Astronomy \\
  The University of Alabama \\
  Tuscaloosa, AL \\
  \texttt{sgleyzer@ua.edu} \\
}

\begin{document}
\maketitle

\begin{abstract}
Recent work has established an alternative to traditional multi-layer perceptron neural networks in the form of Kolmogorov-Arnold Networks (KAN). The general KAN framework uses learnable activation functions on the edges of the computational graph followed by summation on nodes. The learnable edge activation functions in the original implementation are basis spline functions (B-Spline). Here, we present a model in which learnable grids of B-Spline activation functions are replaced by grids of re-weighted sine functions (SineKAN). We evaluate numerical performance of our model on a benchmark vision task. We show that our model can perform better than or comparable to B-Spline KAN models and an alternative KAN implementation based on periodic cosine and sine functions representing a Fourier Series. Further, we show that SineKAN has numerical accuracy that could scale comparably to dense neural networks (DNNs). Compared to the two baseline KAN models, SineKAN achieves a substantial speed increase at all hidden layer sizes, batch sizes, and depths. Current advantage of DNNs due to hardware and software optimizations are discussed along with theoretical scaling. Additionally, properties of SineKAN compared to other KAN implementations and current limitations are also discussed
\end{abstract}

\section{Introduction}
\label{sec:introduction}
Multi-layer perceptrons (MLPs) are a fundamental component of many current leading neural networks \cite{mlp1, mlp2}. They are often combined with feature extracting tools, such as convolutional neural networks \cite{cnn, resnet, gradcomp} and multi-head attention \cite{allyouneed}, to create many of the best performing models, such as transformers. One of the key mechanisms that makes MLPs so powerful is that the layers typically end in non-linear activation functions which enables universal approximation from any arbitrary input space to any arbitrary output space using only a single sufficiently wide layer \cite{mlpuniversal}. While MLPs enable any such arbitrary mapping, the number of neurons required to achieve that mapping can also be arbitrarily large.

Recent work \cite{liu2024kan} has presented an alternative to the MLP architecture, based on the Kolmogorov-Arnold Representation Theorem \cite{Kolmogorov1956, Kolmogorov1957, Braun2009}, accordingly denoted as Kolmogorov-Arnold Networks (KANs) \cite{liu2024kan}. In earlier seminal work, \cite{Kolmogorov1956, Kolmogorov1957}, it was established that any arbitrary multivariate function can be approximated with a sum of continuous univariate functions over a single variable. In \cite{liu2024kan}, it was shown that this approximation can be extrapolated to neural network architectures leading to competitive performance with MLPs at often significantly smaller model sizes \cite{mlp1}\cite{mlp2}. In this work, we will use an efficient implementation of the KAN with learnable B-Spline activation functions (B-SplineKAN) \cite{cao2024efficient-kan} that is numerically consistent with the original implementation of KAN, but on the order of three to five times faster than the original implementation \cite{liu2024kan} for the purpose of performance comparison.

\begin{figure}[!ht]
    \centering
    \includegraphics[width=1.\textwidth]{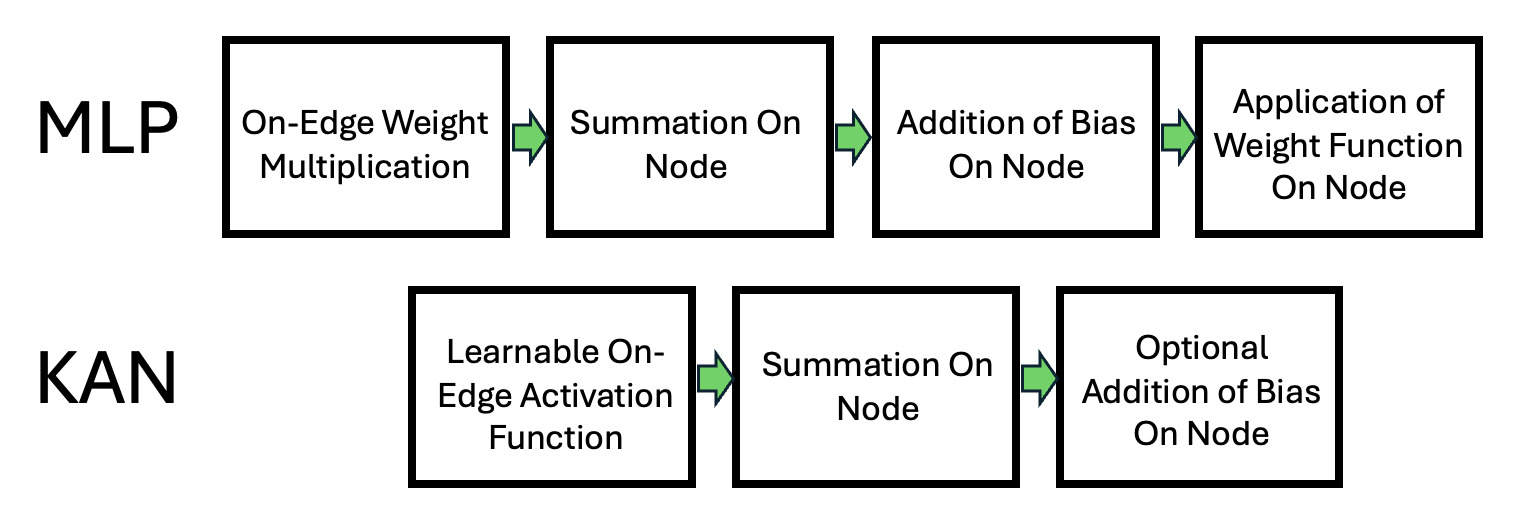}
    \caption{Flow of operations. Top: MLP Bottom: KAN}
    \label{fig:mlpvskan}
\end{figure}

As Figure \ref{fig:mlpvskan} shows, the order of operations in traditional MLPs is: on-edge weight multiplication, summation on node, addition of bias, followed by the application of the activation function. In KANs, the order of operations is: learnable activation function on edge, summation on node and optional addition of bias on node. This alternative order of operations satisfies the Kolmogorov-Arnold Representation Theorem and can potentially allow significantly smaller computational graphs compared to MLPs \cite{liu2024kan}.

The work establishing KAN as a viable model \cite{liu2024kan} explored the use of B-Splines as the learnable activation function. There is a strong motivation for the choice of B-Splines. Using B-Splines, it is possible to change the size of layer's grid of spline coefficients without meaningfully affecting the model itself, enabling downstream fine-tuning. It is also possible to sparsify the model through a process of pruning low-impact spline terms. It is additionally possible to  determine the functional form of the model symbolically. \cite{liu2024kan} found that B-Spline models achieved competitive results with MLP layers on a broad range of tasks. The choice of B-Splines isn't without its costs however, as B-SplineKAN layers are significantly slower than MLPs and, while recent implementations have helped to close the gap, MLPs are still substantially faster. Furthermore, there are many tasks presented in \cite{liu2024kan} where MLPs still outperformed the B-SplineKAN. Recent work has shown that alternatives to B-SplineKAN can achieve competitive performance under fair comparison \cite{fairkancompare}. In this paper we present SineKAN, a KAN implementation with sine activation functions which aims to address size and speed limitations of common KAN models by replacing B-Spline functions with periodic sine functions. We will also compare to an existing periodic KAN model, the FourierKAN \cite{xu2024fourierkangcf}.

In this work we will introduce the novel SineKAN model functional form and provide empirical evidence that it can achieve comparable performance to B-Spline KAN models and outperform FourierKAN models on some common benchmark tasks. We also show that it can partially avoiding catastrophic forgetting during continual learning, a property which has helped drive interest in other KAN models. In Section \cref{sec:sinekan} we describe the SineKAN architecture, whether it satisfies a universal approximation, and outline a weight initialization strategy that scales consistently with differing grid sizes and stabilizes numerical performance across multi-layer models. In Section \cref{sec:results}, we present results of model inference speed and performance on the MNIST benchmark and compare it with B-SplineKAN and FourierKAN implementations. We discuss our results in Section \cref{sec:discussion} and summarize our findings in Section \cref{sec:conclusion}.

\FloatBarrier
\section{Related Work}
A number of alternative univariate functions to B-Splines have been explored for use in KANs, including wavelets \cite{wavkan}, Chebyshev polynomials \cite{chebykan}, fractional functions \cite{fractionalkan}, rational Jacobi functions \cite{rkan}, radial basis functions \cite{rbfkan}, and even variations on Fourier expansions \cite{xu2024fourierkangcf}, discussed in detail in Section \cref{subsec:gridphase}.

Periodic activation functions in neural networks have been explored extensively and shown to provide strong approximations for a broad class of problems. Such problems include general functional modeling \cite{Gallant1988}, image classification \cite{Zhumekenov2019, Wong2002}, and \cite{Sopena1999, Parascandolo2016} for general classification tasks. Work using sinusoidal representational networks \cite{sitzmann2020implicit} has shown that sinusoidal activation functions lead to strong performance on problems with continuous domains \cite{siren3} and potentially discontinuous domains \cite{siren1, siren2}. These promising results in sinusoidal activations motivate sine functions as a potentially strong alternative to other explored activation functions for KANs.

\FloatBarrier
\section{SineKAN}
\label{sec:sinekan}
\subsection{Sinusoidal Activation Function}
Here, we propose an alternative to the B-SplineKAN architecture described in Section \cref{sec:introduction} that is based on sine functions. Mathematically each layer can be expressed as:

\begin{equation}
    y_i = \sum_{j,k}  A_{ijk} \sin(\omega_{k} x_{j} + \phi_{jk}) + b_i
\end{equation}

where $y_i$ are the layer output features, $x_j$ are the layer input features, $\phi_{jk}$ is a phase shift over the grid and input dimensions, $\omega_k$ is a grid frequency, $A_{ijk}$ are the amplitude weights, and $b_i$ is a bias term. The base functional form of the sines are fixed, while the functional form of the sine activations is learned through learnable frequency and amplitude terms performed over a grid of fixed phases.

\subsection{Grid Phase Shift}
\label{subsec:gridphase}
In previous work using Fourier series, KAN networks use the form of a full Fourier series expansion, denoted as the following\cite{xu2024fourierkangcf}:

\begin{equation}
    y_i = \sum_j\sum_k  A_{ijk} \sin(k x_{j}) + B_{ijk}\cos(k x_{j}) + b_i
\end{equation}

where $y_i$ are the layer output features, $x_j$ are the layer input features, $A_{ijk}$ and $B_{ijk}$ are Fourier weight matrices, and $b_i$ is a bias. Here, there is an additional three-dimensional weight matrix compared to SineKAN:

\begin{equation}
    y_i = \sum_{j,k}  A_{ijk} \sin(\omega_{k} x_{j} + \phi_{jk}) + b_i
\end{equation}

By introducing learnable frequencies $\omega_{k}$ over a grid with fixed phase shifts $\phi_{jk}$  and input dimensions, we reduce the number of learnable parameters from $O(2oig)$ to $O(oig+g)$ where o is the output dimension, i is in the input dimension, and g is the grid size. We conjecture later in \cref{subsec:universalapprox} that this still satisfies universal approximation in the large model limit.

\begin{figure}[ht]
    \centering
    \includegraphics[width=1.\textwidth]{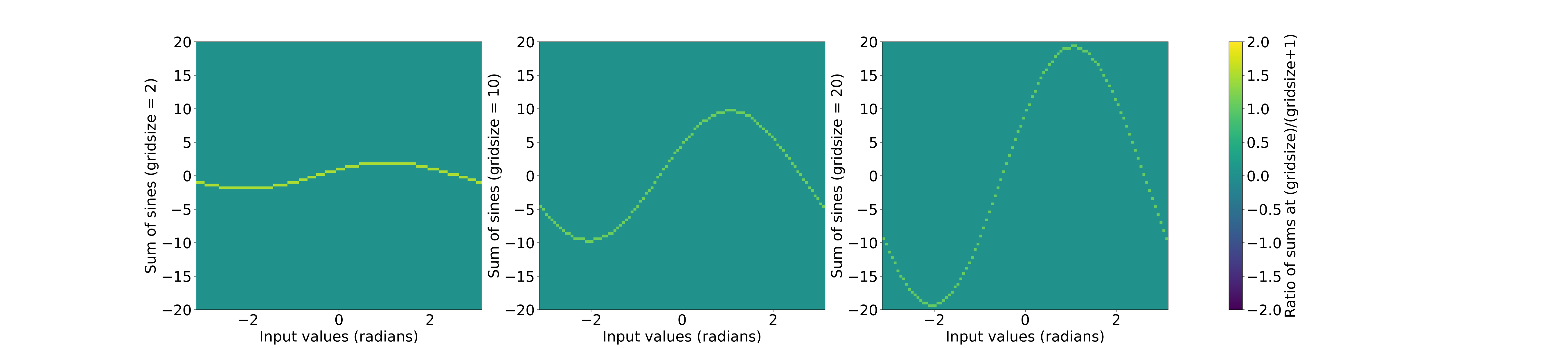}
    \caption{Value of $\sum_{i=1}^g \sin \left(x + \frac{i}{g+1} \right)$ as a function of x with the ratio of sum at g+1 over the sum at g as the color scale. Left to right: $g=2$, $g=10$, $g=20$}
    \label{fig:gridscaling}
\end{figure}

Under the initial assumption for the first layer of the model, a naive approach for initializing the grid weights is to cover a full phase shift range, where the grid phase shift terms would be a range of values from 0 to $\pi$. However, it can be shown that, for the following case: 

\begin{equation}
    \sum_{i=1}^g sin\left(x + \frac{i}{g} \right)
\end{equation}

where g is the grid size, the total sum increases non-linearly as a function of g. Most importantly, the total sum is independent of input value, x. This makes finding the appropriate grid weight scaling inconsistent across types of inputs and grid dimension. We present an alternative strategy, in which grid weights are initialized as: 

\begin{equation}
    \sum_{i=1}^g \sin \left(x + \frac{i}{g+1} \right)
\end{equation}

In the case where frequencies are all fixed at the same constant value, the sum converges to:

\begin{equation}
    \sum_{i=1}^g \sin \left( x + \frac{i}{g+1} \right) = C(g) \sin \left( x + \frac{1}{2} \right)
\end{equation}

where C(g) is a constant that scales with g. This means that, for fixed frequencies, the scaling behavior of the model output would be independent of x. 

Furthermore, we find that introducing an additional input phase term along the axis of the number of input features with values ranging from zero to $\pi$ leads to stronger model performance.

Finally, to stabilize the model scaling across different grid sizes, we find a functional form that helps scale the total sum across the grid dimension as a ratio of phase terms: 

\begin{equation}
    R = A g^{-K} + C    
    \phi_{g+1} = \phi_{g} R(g)
\end{equation}

where $A = 0.97241$, $K = 0.988440$ and $C = 0.999450$, R is a scale factor by which all phase terms are multiplied as you increase from a grid size of one upward, and $\phi_g$ is the phase at a particular grid size. To determine A, K, and C we perform least squares minimization of: 

\begin{equation}
    L(g,x) = \sigma^2 \left( \frac{f(g+1,x)}{f(g,x)} \right) + \left( 1 - \mu\left( \frac{f(g+1,x)}{f(g,x)} \right) \right)^2
\end{equation}

where L is a cost function, f(g,x) is the sum of sines across input values from $-\pi$ to $\pi$, $\mu$ is the mean value and $\sigma^2$ is the standard deviation.

\begin{figure}[ht]
    \centering
    \includegraphics[width=1.\textwidth]{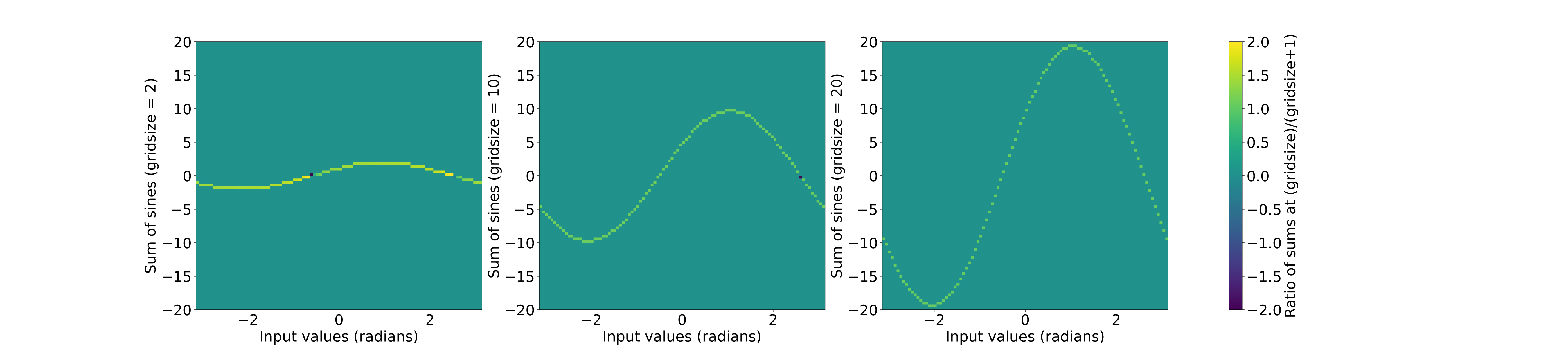}
    \caption{$\sum_{k=1}^g \sin\left(x + \frac{k\pi}{g+1} R(g)\right)$ with the ratio of sum at $g+1$ over the sum at g as the color scale. Left to right: $g=2$, $g=10$, $g=20$}
    \label{fig:gridscalingperturbed}
\end{figure}

Using this functional form we can initialize layers with arbitrary grid sizes by using the exponential expression in a recursive formula while initializing the grid phase weights. The resulting ratios of sums of sines are shown in Figure \ref{fig:gridscalingperturbed}.

\begin{figure}[!ht]
    \centering
    \includegraphics[width=1.\textwidth]{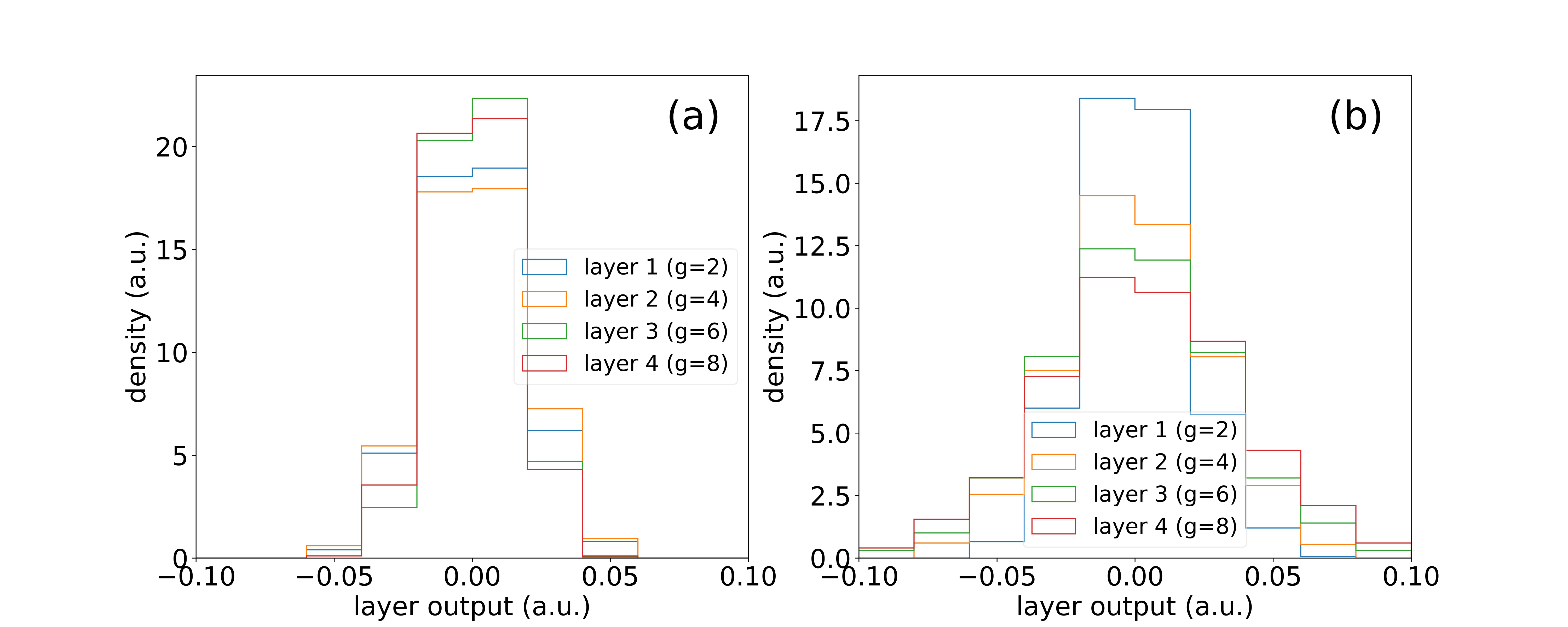}
    \caption{(a) Outputs of layers of same size (N=1000) with the recursive function applied for grid size scaling. (b) Outputs of layers of same size (N=1000) without the recursive function applied for grid size scaling.}
    \label{fig:gridsizescaling}
\end{figure}

In Figure \ref{fig:gridsizescaling}(b), we see the outputs of subsequently connected layers when recursive grid size phase scaling is not applied. In Figure \ref{fig:gridsizescaling}(a), we see the same scenario but with recursive grid size phase scaling applied and see an increase in similarity of layer outputs across various grid sizes.

\FloatBarrier
\subsection{Scaling of Phase Terms With Grid Size}
\label{subsec:fourierweights}
\begin{figure}[!ht]
    \centering
    \includegraphics[width=\textwidth]{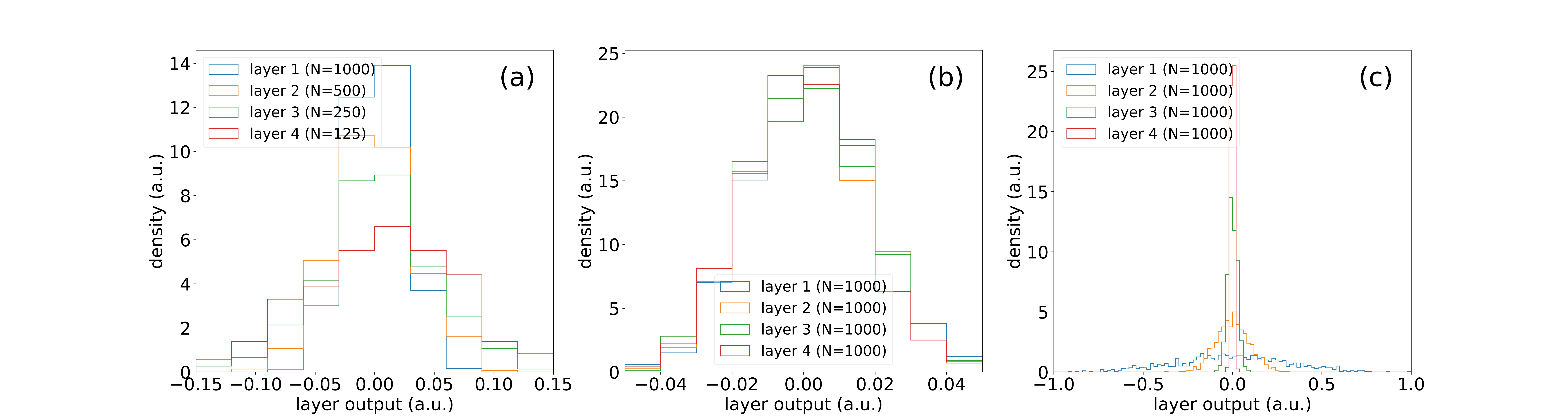}
    \caption{(a) Outputs of consecutive layers of different sizes in a SineKAN model. (b) Outputs of consecutive layers of same size in a SineKAN model. (c) Outputs of consecutive layers of same size in a B-SplineKAN model.}
    \label{fig:layeroutput}
\end{figure}

We find a weight initialization strategy which results in strong performance and stability in higher depth models. For the first layer, the weights are initialized as a random distribution with a mean of 0 and and a standard deviation of 0.4 and for the subsequent layers, the weights are initialized from a uniform distribution between -1 and 1. This not only leads to consistent layer weight distributions, but also leads to consistent output across connected layers of same size, as shown in Figure \ref{fig:layeroutput}(b). 

It also features a desirable property that, given a fixed initialization on the first layer and a random input vector, the features span a broader range of values at deeper layers, as shown in Figure \ref{fig:layeroutput}(a). This implies that no value collapse is introduced. Comparatively, we see in Figure \ref{fig:layeroutput}(c), that the model layer outputs in B-SplineKAN implementations decrease in multi-layer models, which can play a significant role in results in Section \cref{subsec:inferenceperformance}.

\FloatBarrier
\subsection{Universal Approximation}
\label{subsec:universalapprox}
The combination of learnable amplitudes and sinusoidal activation functions have previously been shown to be viable for implicit neural representations \cite{sitzmann2020implicit}. The models have been shown to be effective for applications including control systems \cite{siren1}, medical applications \cite{siren2}, and physics applications \cite{siren3}. However, these models only satisfy universal approximation on a single layer when combined with linear transformations in sufficiently wide or deep models. By introducing an additional degree of freedom in the form of learnable frequencies over phase-shifted grids, one can eliminate the linear transformation layers.

Any sufficiently well-behaved/smooth 1 dimensional function $f:\mathbb{R}\to\mathbb{R}$
can be expressed in terms of a \emph{Fourier transform} $\tilde{f}:\mathbb{R}\to\mathbb{C}$
w.r.t. a continuous phase space of frequencies $\omega$:
\begin{align*}
f(x) & =\int_{\mathbb{R}}d\omega\;\tilde{f}(\omega)e^{i\omega x}\\
 & =\mathsf{Re}\left(\int_{\mathbb{R}}d\omega\;\tilde{f}(\omega)e^{i\omega x}\right)\\
 & =\int_{\mathbb{R}}d\omega\;A(\omega)\cos\left(\omega x+\phi'(\omega)\right)\\
 & =\int_{\mathbb{R}}d\omega\;A(\omega)\sin\left(\omega x+\phi(\omega)\right)
\end{align*}
where $A(\omega)$ and $\phi(\omega)=\phi'(\omega)-\frac{\pi}{2}$
are real-valued functions. The above integral can be discretized using
Riemannian sums over a finite set of frequencies $\Omega=\left\{ \omega_{1},\omega_{2}\dots\omega_{G}\right\} $
where cardinality $G$ of the set is the \emph{grid size}. We henceforth
propose the following variational function $g_{\theta}$ as an ansatz
for $f(x)$:
\[
g_{\theta}(x)=\sum_{i}B_{i}\sin\left(\omega_{i}x+\phi_{i}\right)
\]
where we make the replacements $\int_{\mathbb{R}}\to\sum_{i}$, $d\omega A(\omega)\to B_{i}$ and $\omega,\phi(\omega)\to\omega_{i},\phi_{i}$. Here, $\phi_{i}$ are G fixed, finite points from (0,$\pi+1$] while we treat all other subscripted symbols $B_{i},\omega_{i}$ as weights whose values can trained to optimize a loss function between $f$ and $g_{\theta}$, which converges to the Fourier transform integral of $f$ as $G\to\infty$. Hence, in the limit where $G\to\infty$, it's a valid candidate for a learnable activation function ansatz to be used in a Kolmogorov-Arnold network (KAN).

\begin{figure}[!ht]
    \centering
    \includegraphics[width=\textwidth]{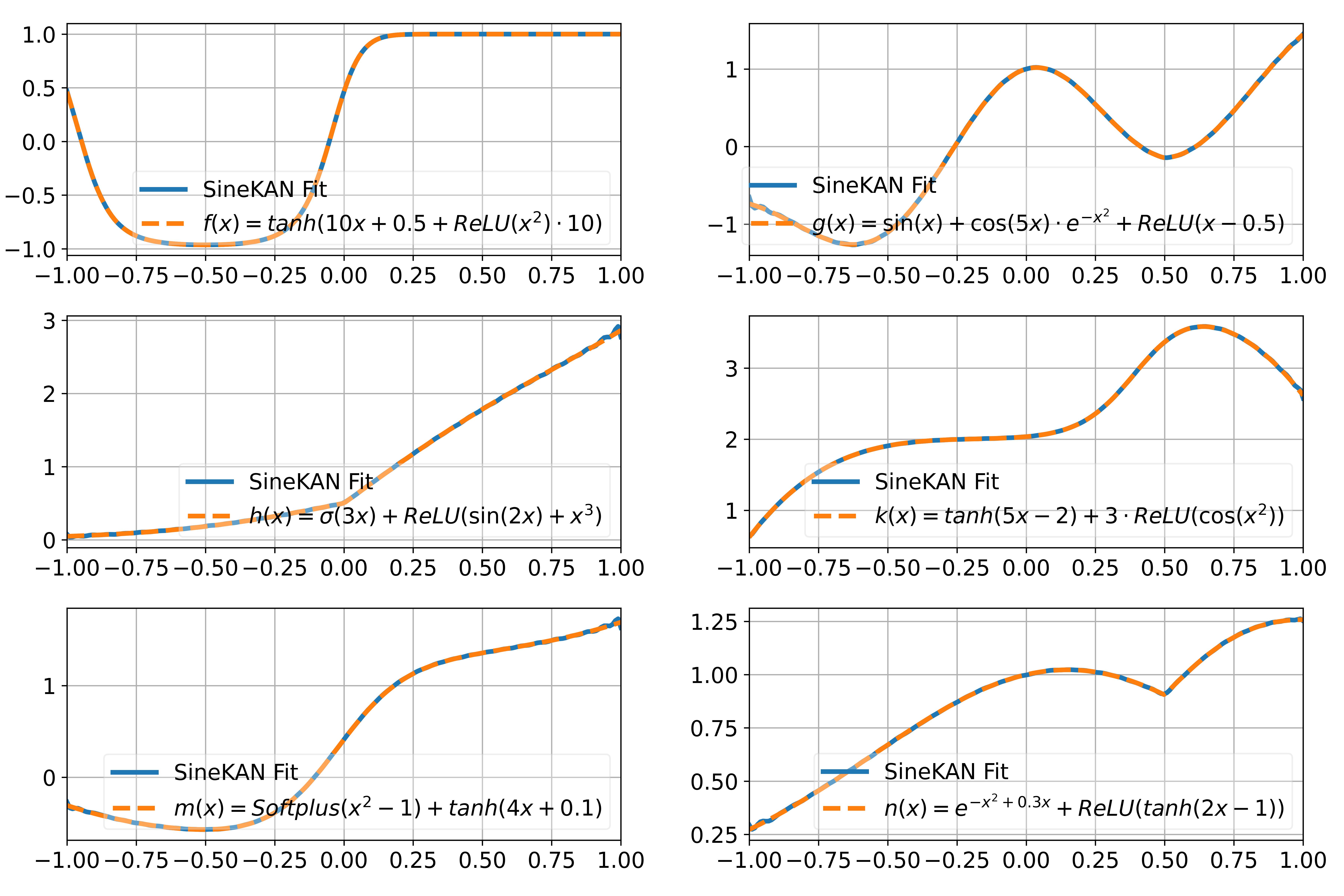}
    \caption{Fits of SineKAN with grid size of 100 to assorted functions.}
    \label{fig:universalapprox}
\end{figure}

In \ref{fig:universalapprox} we show that with a layer with grid size of 100, a single-input, single-output SineKAN layer can map inputs to outputs in 1D functions including over non-smooth functional mappings. The functions explored are:

\begin{itemize}
 \item $f(x) = \tanh(10x + 0.5 + \text{ReLU}(x^2) \cdot 10)$
 \item $g(x) = \sin(x) + \cos(5x) \cdot e^{-x^2} + \text{ReLU}(x - 0.5)$ 
 \item $h(x) = \sigma(3x) + \text{ReLU}(\sin(2x) + x^3)$
 \item $k(x) = \tanh(5x - 2) + 3 \cdot \text{ReLU}(\cos(x^2))$
 \item $m(x) = \text{Softplus}(x^2 - 1) + \tanh(4x + 0.1)$
 \item $n(x) = e^{-x^2 + 0.3x} + \text{ReLU}(\tanh(2x - 1))$
\end{itemize}

\FloatBarrier
\section{Results}
\label{sec:results}
\subsection{Continual Learning}
\label{subsec:continuallearning}

\begin{figure}[!ht]
    \centering
    \includegraphics[width=.45\textwidth]{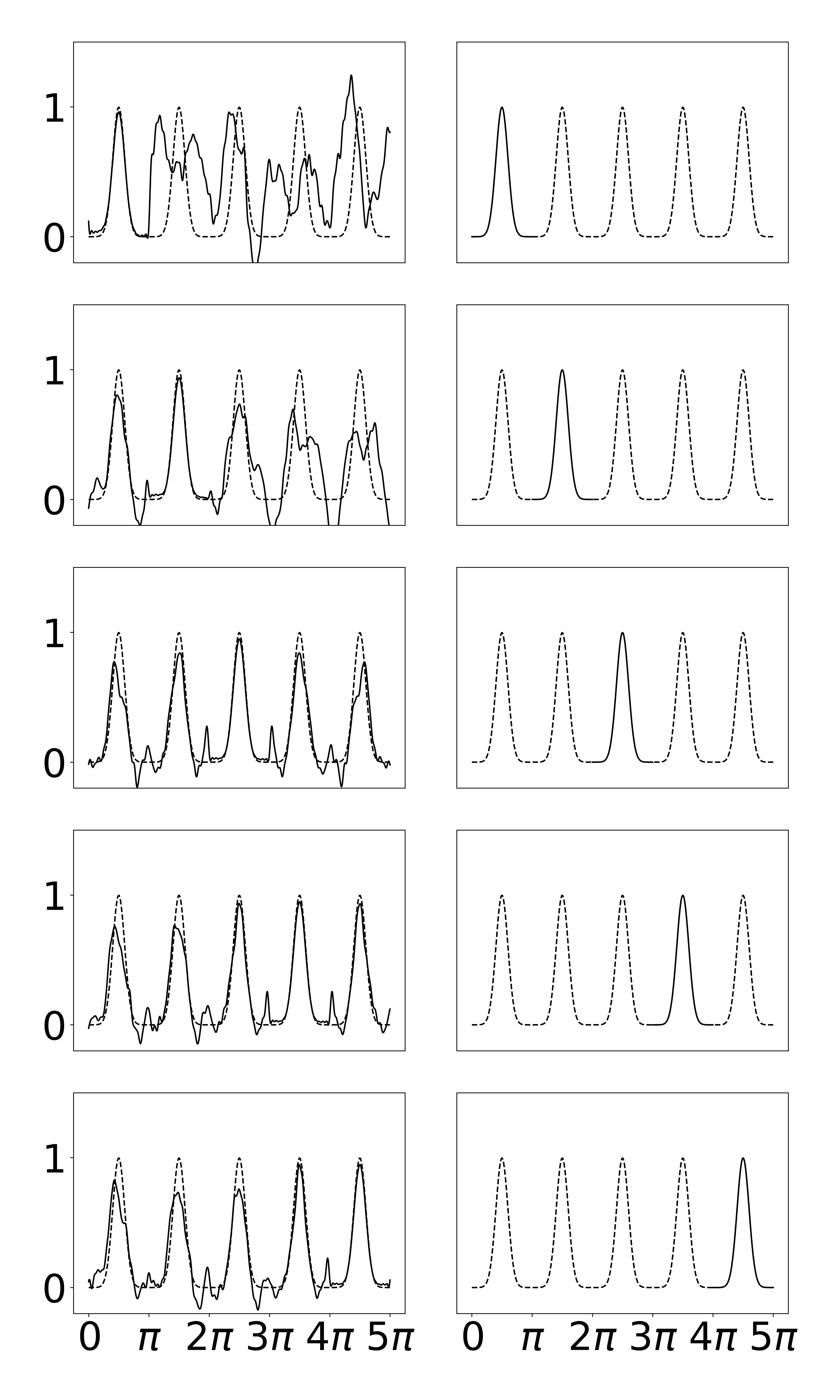}
    \caption{Left: SineKAN model predictions across entire domain; Right: Portion of domain shown to model during progressive training}
    \label{fig:contlearn}
\end{figure}

A desirable property demonstrated for B-SplineKAN models is the ability to fit to sections of the total domain without "forgetting" other sections. In \cite{liu2024kan}, this was performed by fitting a repeating Gaussian waves one period at a time. The first period is fit then the first period is removed from the training data and replaced with the second period. This is repeated for each of the five periods. The original B-SplineKAN work showed that the model could learn to fit data in the new portion of the domain without catastrophically forgetting the other portions of the data it had been shown.

This task presents challenges for SineKAN due to the fact that the range of the function is periodically repeating even beyond the subsection of the domain where the model is initially fit. Due to this we expect SineKAN to have greater difficulty avoiding some amount of "forgetting". However, because the model is periodic, another behavior can emerge in which the model, if able to generalize the pattern across subsections, can potentially forecast into regions of the domain it hasn't been exposed to. This is a potentially very desirable property for tasks where there is any kind of periodic symmetry across the domain. 

\begin{figure}[!ht]
    \centering
    \includegraphics[width=.45\textwidth]{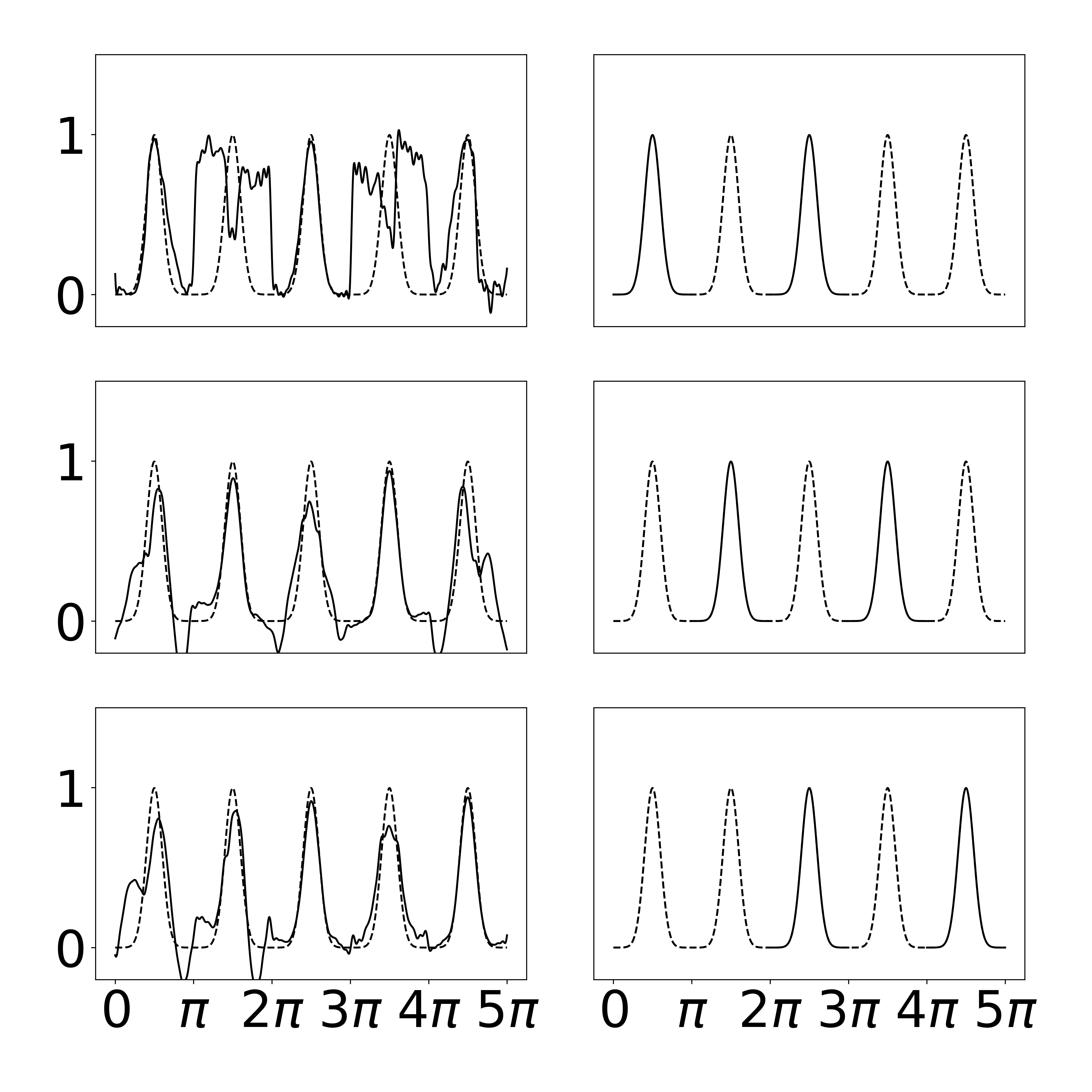}
    \caption{Left: SineKAN model predictions across entire domain; Right: Portion of domain shown to model during progressive training}
    \label{fig:contlearnsp}
\end{figure}

We see in \cref{fig:contlearn} that the SineKAN model exhibits the expected behavior of experiencing some amount of forgetting. We also see that by the second period it begins to generalize the periodic behavior to the third period and, by the third period, it's generally able to capture the period behavior across the entire domain. In \cref{fig:contlearnsp}, we show that exposing the model to more that one disconnected period at a time (period 1 and 3 then 2 and 4 then 3 and 5) also allows the model to capture the periodic behavior across the entire domain with twice as many forward passes but with the same number of total back propagation steps. This implies that discontinuous domains wouldn't necessarily limit the model's generalization behavior.

\FloatBarrier
\subsection{KAN Numerical Performance on MNIST}
\label{subsec:inferenceperformance}
The MNIST dataset is a classification dataset which contains 60,000 training examples and 10,000 testing examples of handwritten characters \cite{deng2012mnist}. The characters can have values between 0 and 9. We train and compare single-layer B-SplineKAN and SineKAN networks on the MNIST dataset. We use a batch size 128, learning decay rate of 0.9 and learning rates of 5e-3 for B-Spline, 1e-4 for FourierKAN, and 4e-4 for SineKAN, optimized with grid search. The models are trained using the AdamW optimizer with a weight decay of 0.01 for B-SplineKAN, 1 for FourierKAN, and 0.5 for SineKAN also found via grid search \cite{loshchilov2019decoupledweightdecayregularization}. We test model performance with single hidden layer dimensions of 16, 32, 64, 128, 256, 512 training for 30 epochs using cross entropy loss \cite{rumelhart1986learning}.

\begin{table}[h!]
\centering
\begin{tabular}{|l|c|c|c|c|c|}
\hline\hline
Layer Size & Model & Accuracy & Precision & Recall & F1 \\
\hline\hline
16 & Sine & \textbf{0.9616} & \textbf{0.9617} & \textbf{0.9616} & \textbf{0.9616}\\
16 & B-Spline & 0.9568 & 0.9571 & 0.9568 & 0.9568\\
16 & Fourier & 0.9337 & 0.9337 & 0.9337 & 0.9336\\
\hline
32 & Sine & \textbf{0.9766} & \textbf{0.9766} & \textbf{0.9766} & \textbf{0.9766}\\
32 & B-Spline & 0.9711 & 0.9711 & 0.9711 & 0.9711\\
32 & Fourier & 0.9499 & 0.9500 & 0.9499 & 0.9499\\
\hline
64 & Sine & \textbf{0.9818} & \textbf{0.9818} & \textbf{0.9818} & \textbf{0.9818}\\
64 & B-Spline & 0.9790 & 0.9790 & 0.9790 & 0.9790\\
64 & Fourier & 0.9597 & 0.9597 & 0.9597 & 0.9596\\
\hline
128 & Sine & \textbf{0.9850} & \textbf{0.9850} & \textbf{0.9850} & \textbf{0.9850}\\
128 & B-Spline & 0.9802 & 0.9802 & 0.9802 & 0.9802\\
128 & Fourier & 0.9656 & 0.9656 & 0.9656 & 0.9656\\
\hline
256 & Sine & \textbf{0.9853} & \textbf{0.9853} & \textbf{0.9853} & \textbf{0.9853}\\
256 & B-Spline & 0.9834 & 0.9834 & 0.9834 & 0.9834\\
256 & Fourier & 0.9709 & 0.9709 & 0.9709 & 0.9709\\
\hline\hline
\end{tabular}
\caption{Model performance metrics by layer size with the best scores bolded.}
\label{tab:kan_comparison}
\end{table}

\begin{figure}[!ht]
    \centering
    \includegraphics[width=1.\textwidth]{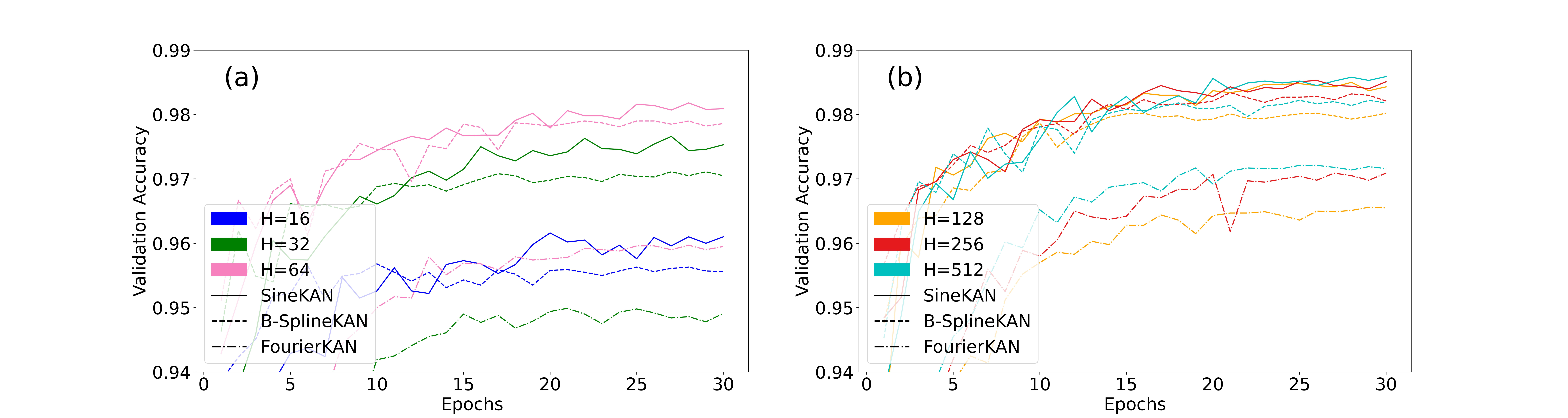}
    \caption{Validation accuracy of B-SplineKAN, FourierKAN, and SineKAN on MNIST with a single hidden layer of size (A) 16, 32, and 64 and (B) 128, 256, and 512.}
    \label{fig:valacc}
\end{figure}

Fig. \ref{fig:valacc} shows the model validation accuracy as a function of the number of epochs. The best accuracies are shown in Table \ref{tab:kan_comparison}. The SineKAN model achieves better results than the FourierKAN and B-SplineKAN models for all hidden layer sizes.

\begin{figure}[!ht]
    \centering
    \includegraphics[width=.65\textwidth]{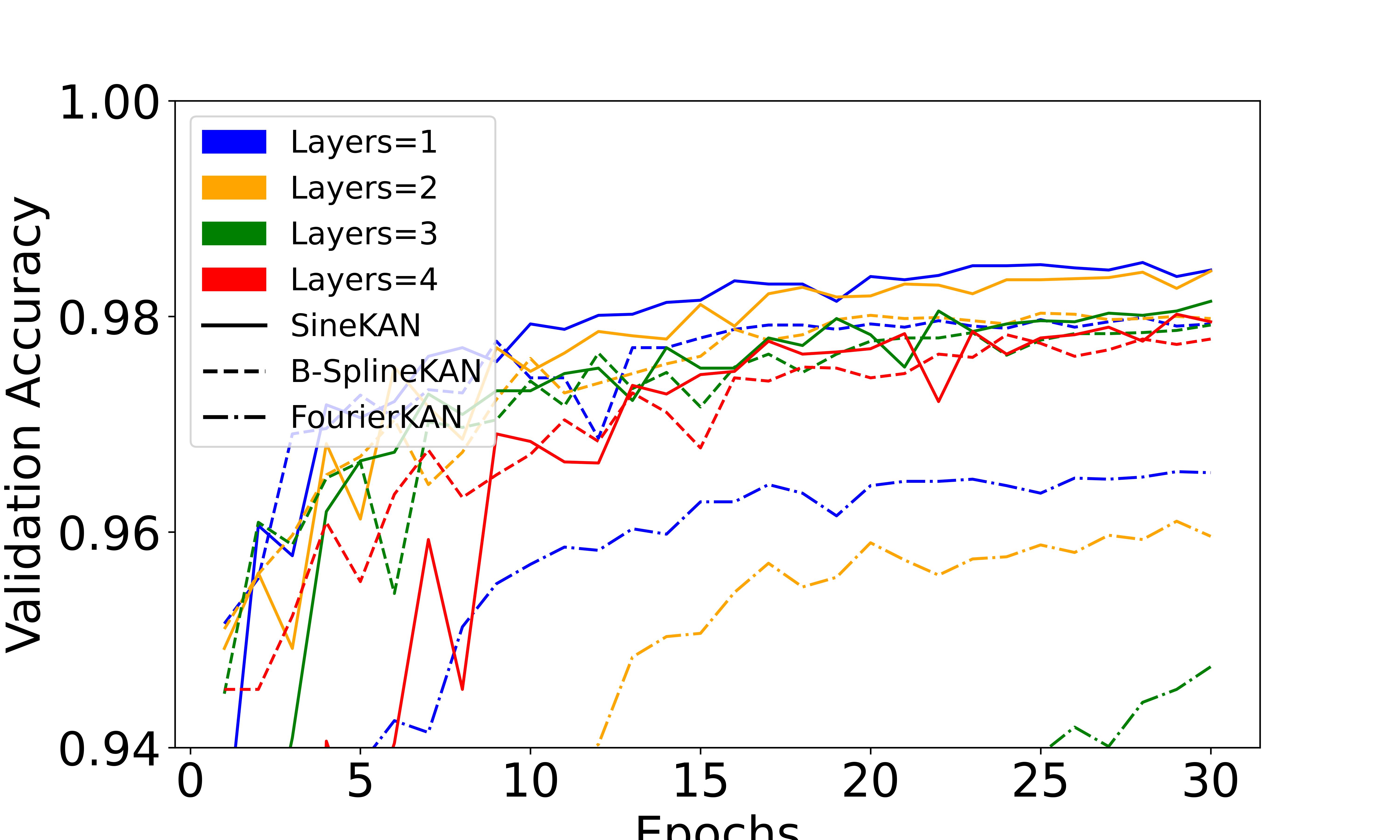}
    \caption{B-SplineKAN, FourierKAN, and SineKAN validation accuracy on MNIST with a 1, 2, 3, and 4 hidden layers of size 128.}
    \label{fig:valacclayers}
\end{figure}

We additionally explore fitting using the same hyperparameters but with 1, 2, 3, and 4 hidden layers of size 128. Figure \ref{fig:valacclayers} shows that the SineKAN outperforms the FourierKAN and B-SplineKAN at lower layer depths. We also find, however, that without additional tuning of hyperparameters, all three models generally decrease in performance at higher depths.

\FloatBarrier
\subsection{KAN Inference Speeds}
\label{subsec:inferencespeed}
We benchmark the speed of SineKan and B-SplineKAN models using NVIDIA T4 GPU with 16GB of RAM. We test performance on variable batch sizes of 16, 32, 64, 128, 256, 512 on single inputs of 784 features using a single hidden layer of size 128 with a grid size of 8. We test performance on single hidden layer hidden dimensions of 16, 32, 64, 128, 256, 512 under the same conditions. We test performance with single batch of 784 features with 1, 2, 3, and 4 hidden layers of size 128.

\begin{figure}[!ht]
    \centering
    \includegraphics[width=1.\textwidth]{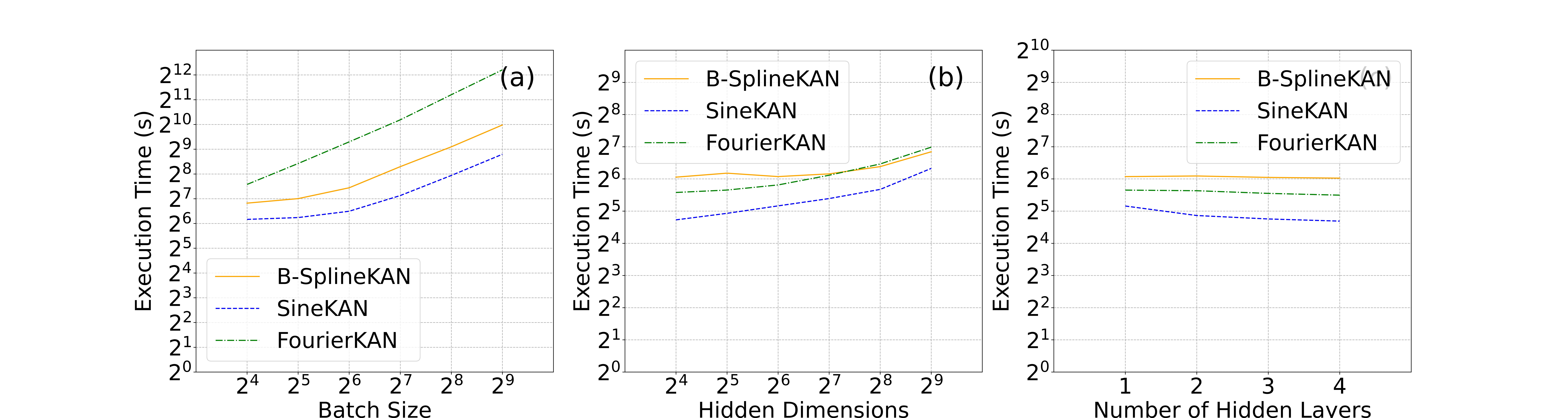}
    \caption{Average inference times (averaged over 1,000 passes) as a function of (a) batch size, (b) hidden dimension, (c) hidden layers of B-SplineKAN, FourierKAN, and SineKAN run on NVIDIA Tesla T4 GPU (16GB RAM).}
    \label{fig:scaling}
\end{figure}

As Figure \ref{fig:scaling} shows, the SineKAN has the best inference times at all batch sizes compared to B-SplineKAN and FourierKAN. It also has the best inference times across all hidden layer dimensions explored with all three models showing roughly flat scaling as a function of model depth. Due to differences in hardware optimization and cuda kernel optimization, it is difficult to directly compare the performance of the three models on-device. To account for idealized performance optimizations, we also derive analytically the expected model scaling behavior in \cref{subsec:model_scaling}. We find that the expected scaling for the three models in approximate FLOP compute units are as follows:
\begin{itemize}
    \item SineKAN: $\mathcal{O}(2 b \; d_{out} \; d_{in} \; g)$
    \item FourierKAN: $\mathcal{O}(4 b\;d_{out} \; d_{in} \; g)$
    \item B-SplineKAN: $\mathcal{O}(2 b\; d_{out}\; d_{in}\; (g + s))$
\end{itemize}
Here $b$ is the batch size, $d_{in}$ and $d_{out}$ are the input and output dimensions, $g$ is the grid size, and $s$ is the basis-spline order. We therefore expect that, with full device-level and software-level optimization, the relative performance of B-SplineKAN will scale like $g/(g+s)$ relative to SineKAN and FourierKAN to scale like half the speed of SineKAN.

\subsection{MLP Comparison}
We also compare performance to MLP in \cref{tab:mlp_metrics}. For MLP we performed a similar grid search and found a learning rate of 8e-4, weight decay of 0.01, and learning decay rate of 0.9 to be the best performing. We find that MLP exceeds SineKAN's performance at a grid size of 8 and hidden layer size of 256 once it reaches a hidden layer size of 2048.

\begin{table}[h!]
\centering
\begin{tabular}{|l|c|c|c|c|c|}
\hline\hline
Layer Size & Model & Accuracy & Precision & Recall & F1 \\
\hline\hline
16 & MLP & 0.9183 & 0.9181 & 0.9183 & 0.9181\\
32 & MLP & 0.9557 & 0.9557 & 0.9557 & 0.9556\\
64 & MLP & 0.9682 & 0.9682 & 0.9682 & 0.9682\\
128 & MLP & 0.9797 & 0.9797 & 0.9797 & 0.9797\\
256 & MLP & 0.9822 & 0.9822 & 0.9822 & 0.9822\\
512 & MLP & 0.9842 & 0.9842 & 0.9842 & 0.9842\\
1024 & MLP & 0.9843 & 0.9843 & 0.9843 & 0.9843\\
2048 & MLP & 0.9863 & 0.9863 & 0.9863 & 0.9863\\
\hline
\end{tabular}
\caption{MLP performance metrics by layer size.}
\label{tab:mlp_metrics}
\end{table}

Due to previously mentioned differences in device-level and software-level optimizations for the different models, it's difficult to directly compare the performance of MLP and SineKAN. However, the characteristic FLOPs of an MLP layer scales as $\mathcal O \left(2 b d_{out} d_{in} \right)$. We therefore would consider that, for a SineKAN model with a grid size of 8, a competitive MLP performance would be at 8 times the hidden layer dimension. However, it's also worth acknowledging that as the first hidden layer size in the MLP increases, additional parameters are added in the output layer. MLP under-performing SineKAN (0.9853 accuracy) at a hidden layer size of 1024 (0.9843 accuracy) and outperforming SineKAN at a hidden layer size of 2048 (0.9863 accuracy) is consistent with a competitive performance to MLP as a function of FLOPs under full optimization.

\FloatBarrier
\section{Discussion}
\label{sec:discussion}

The SineKAN model, which uses sine functions as an alternative to existing baseline B-Spline activation functions \cite{liu2024kan}, shows very good performance on the benchmark task. Model stabilization techniques described in Section \cref{subsec:gridphase} lead to consistent model layer output weights at different depths and sizes of phase shift grids. We also show that it actively mitigates value collapse in deep models.

In Section \cref{subsec:inferenceperformance}, we show that SineKAN increasingly outperforms the B-SplineKAN model at very large hidden layer dimensions. These results suggest that the SineKAN model may be a better choice compared to B-SplineKAN for scalable, high-depth, large batch models such as large language models (LLMs). However, when comparing to MLP we also account for difference in scaling resulting from lack of a grid. We find that MLP models perform similarly to SineKAN when accounting for idealized model scaling. However, at very large sizes (hidden dimension of 2048), MLP outperforms SineKAN with comparable idealized inference time.

We show in Section \cref{subsec:inferencespeed} that the SineKAN model is faster than both FourierKAN and B-SplineKAN as a function of batch size, hidden layer dimension, and depth. Due to device-level and software-level optimizations to computation, we also derive empirically in \cref{subsec:model_scaling} what the expected true scaling is under optimal conditions. We find that SineKAN is roughly (g+s)/g times faster than B-SplineKAN where g is grid size and s is basis-spline order and roughly two times faster than FourierKAN.

We also found that SineKAN had a significantly different optimal learning rate and weight decay compared to B-SplineKAN and FourierKAN which motivates the idea that a fair comparison cannot be done across different KAN implementations without performing a grid search to find optimal hyperparameters. Further, we also showed in Section \cref{subsec:fourierweights}, that B-SplineKAN has an inherent flaw in scaling to multi-layer models in that it increasingly constricts layer outputs at higher depths. This likely necessitates additional layer normalization between B-SplineKAN layers. We recommend that approaches for stabilizing model weights at different sizes and depths, similar to those outlined in \cref{subsec:fourierweights} and \cref{subsec:gridphase}, should be employed in other KAN models to improve deep model stability and performance.

Regarding general KAN properties, SineKAN is able to demonstrate some avoidance of catastrophic forgetting in the continual learning task. It also shows a potentially favorable behavior in generalization of repeating patterns to as-of-yet unseen portions of the domain space. However, we introduce a recursive phase scaling which improves model stability and performance at different model sizes. This makes the current implementation of SineKAN incapable of directly transferring weights to a larger grid. Absence of grid expandability could present major limitations in use cases where extremely large grid sizes might be required. Further, SineKAN currently does not support symbolic expressions.

In summary, sinusoidal activation functions appear to be a promising candidate in the development of Kolmogorov-Arnold models. SineKAN has superior performance in inference speed and accuracy, as well as multi-layer scaling when compared with B-SplineKAN. However, a number of other activation functions mentioned in Section \cref{sec:introduction} have also shown to have superior inference speed and better numerical performance. Further exploration is needed to compare the performance both in terms of inference speed and numerical performance on the broad range of KAN implementations, and on a broader range of tasks, under fair conditions.

\FloatBarrier
\section{Conclusion}
\label{sec:conclusion}
We present the SineKAN model, a sinusoidal activation function alternative to B-SplineKAN and FourierKAN Kolmogorov-Arnold Networks and multi-layer perceptrons. We find that SineKAN has one desirable property of KAN models in avoidance of catastrophic forgetting during continual learning and an additional property of generalization of patterns to unseen regions of the domain. We also find that this model leads to better numerical performance on the MNIST benchmark task compared to other KAN models and comparable performance to MLP when all models are trained using near-optimal hyperparameters found with a parameter grid search. The SineKAN model outperforms B-SplineKAN at higher hidden dimension sizes with more predictable performance scaling at higher depths. We further find that SineKAN outperforms efficient implementations of FourierKAN and B-SplineKAN on speed benchmarks and are expected to outperform even at full device- and software-level optimization. We find that SineKAN performs similarly to MLP when accounting for idealized scaling though MLP can still outperform SineKAN given sufficiently large hidden layer dimensions.

Future work should aim to compare other KAN models under similar, optimized conditions. Additional explorations are also needed regarding deep-model stabilization techniques for various KAN models. Due to competitive performance of SineKAN models with MLP, we find it worth exploring use of SineKAN models in place of MLP in more complex architectures involving feature extractors such as convolutional neural networks \cite{resnet} and transformers \cite{allyouneed}. Further work is also needed to include features in SineKAN models which are available in some other KAN models \cite{liu2024kan} such as symbolic equation representing and transfer of weights during grid size expansion of the model. Finally, further exploration is needed to determine use cases which best leverage the periodic behavior of the SineKAN model.

\section{Appendix}
\label{sec:appendix}
\subsection{Code}
\label{subsec:code}
The SineKAN code can be found at \href{https://github.com/ereinha/SineKAN}{https://github.com/ereinha/SineKAN}.
\subsection{Model scaling derivations}
\label{subsec:model_scaling}
For these derivations we assume that addition, multiplication, and subtraction will require on average 1 FLOP, division and exponential will require on average 5 FLOPs, and trigonometric functions (sine/cosine) will require on average 10 FLOPs. We will also assume any boolean logic and reshaping will require 0 FLOPs. Here b is the batch size, g is the grid size, s is the basis-spline order, $d_{in}$ is the input dimension, $d_{out}$ is the output dimension.

\textbf{SineKAN Layer:}\\
Initial multiplication:
\begin{equation}
    M_1 = b d_{in} g
\end{equation}
Add phase:
\begin{equation}
    A_1 = b d_{in} g
\end{equation}
Trigonometric evaluations:
\begin{equation}
    N_{sin} = 10 b d_{in} g
\end{equation}
Einsum multiplications:
\begin{equation}
    M_2 = b d_{out} d_{in} g
\end{equation}
Einsum additions:
\begin{equation}
    A_2 = b d_{out} (d_{in} g - 1)
\end{equation}
Add bias:
\begin{equation}
    A_3 = b d_{out}
\end{equation}
Total FLOPs: $b d_{in} g (2 d_{out} + 12) + b d_{out}$ \\
Leading order: $O(2 b d_{in} g d_{out})$

\textbf{FourierKAN Layer:}\\
Initial multiplication:
\begin{equation}
    M_1 = b d_{in} g
\end{equation}
Trigonometric evaluations:
\begin{equation}
    N_{cos} = N_{sin} = 10 b d_{in} g
\end{equation}
Multiplication by Fourier coefficients:
\begin{equation}
    M_2 = 2 b d_{out} d_{in} g
\end{equation}
Addition over dimensions:
\begin{equation}
    A_1 = 2 b d_{out} (d_{in} g - 1)
\end{equation}
Cosine and sine combination:
\begin{equation}
    A_2 = b d_{out}
\end{equation}
Add bias:
\begin{equation}
    A_3 = b d_{out}
\end{equation}
Total FLOPs: $b d_{in} g (4 d_{out} + 21) + b d_{out}$ \\
Leading order: $O(4 b d_{in} g d_{out})$

\textbf{EfficientKAN Layer:}\\
SiLU activation:
\begin{equation}
    N_{silu} = 13 b d_{in}
\end{equation}
Base linear multiplication:
\begin{equation}
    M_{base} = b d_{out} d_{in}
\end{equation}
Base linear addition:
\begin{equation}
    A_{base} = b d_{out} (d_{in} - 1)
\end{equation}
B-Spline basis calculation:
\begin{equation}
    N_{spline} = 17 s b d_{in} (g + 2 s)
\end{equation}
Spline linear multiplication:
\begin{equation}
    M_{spline} = b d_{out} d_{in} (g + s)
\end{equation}
Spline linear addition:
\begin{equation}
    M_{spline} = b d_{out} [d_{in} (g + s) - 1]
\end{equation}
Combine outputs:
\begin{equation}
    A_{combine} = b d_{out}
\end{equation}
Total FLOPs: $13 b d_{in} + 2 b d_{out} d_{in} + 17 s b d_{in} (g + 2s) + 2 b d_{out} d_{in} (g + s)) + b d_{out}$ \\
Leading order: $O(2 b d_{out} d_{in} (g + s))$

\FloatBarrier
\section{Acknowledgements}
This work was supported by the U.S. Department of
Energy (DOE) under Award No. DE-SC0012447 (E.R. and S.G.). E.R. was a participant in the 2023 Google Summer of Code Program.

\bibliographystyle{unsrt}  
\bibliography{references}

\end{document}